\documentclass[sigconf,fontsize=9pt]{acmart}

\usepackage{booktabs} 
\usepackage[utf8]{inputenc} 
\usepackage[T1]{fontenc}    
\usepackage{lmodern}
\usepackage{hyperref}       
\usepackage{url}            
\usepackage{booktabs}       
\usepackage{amsfonts}       
\usepackage{nicefrac}       
\usepackage{microtype}      
\usepackage{graphicx}
\usepackage{wrapfig}
\usepackage{enumitem}
\usepackage{amsmath}
\usepackage{amssymb}
\usepackage{multirow}
\usepackage{makecell}
\usepackage{xcolor, colortbl}

\acmConference[DAC'19]{ACM/IEEE Design Automation Conference}{June 2019}{Las Vegas, Nevada, USA}
\acmYear{2019}
\copyrightyear{2019}
\renewcommand\footnotetextcopyrightpermission[1]{} 
\definecolor{Gray}{gray}{0.85}
\newcolumntype{a}{>{\columncolor{Gray}}c}
\begin{document}
\title{Resource-Scalable CNN Synthesis for IoT Applications}
\author{Mohammad Motamedi}
\affiliation{%
  \institution{University of California, Davis}
  \city{Davis}
  \state{California}
  \postcode{95616}
}

\author{Felix Portillo}
\affiliation{%
	\institution{University of California, Davis}
	\city{Davis}
	\state{California}
	\postcode{95616}
}
\author{Mahya Saffarpour}
\affiliation{%
	\institution{University of California, Davis}
	\city{Davis}
	\state{California}
	\postcode{95616}
}
\author{Daniel Fong}
\affiliation{%
	\institution{University of California, Davis}
	\city{Davis}
	\state{California}
	\postcode{95616}
}
\author{Soheil Ghiasi}
\affiliation{%
	\institution{University of California, Davis}
	\city{Davis}
	\state{California}
	\postcode{95616}
}

\begin{abstract}
State-of-the-art image recognition systems use sophisticated Convolutional Neural Networks (CNNs) that are designed and trained to identify numerous object classes. Such networks are fairly resource intensive to compute, prohibiting their deployment on resource-constrained embedded platforms. On one hand, the ability to classify an exhaustive list of categories is excessive for the demands of most IoT applications. On the other hand, designing a new custom-designed CNN for each new IoT application is impractical, due to the inherent difficulty in developing competitive models and time-to-market pressure. To address this problem, we investigate the question of: ``Can one utilize an existing optimized CNN model to automatically build a competitive CNN for an IoT application whose objects of interest are a fraction of categories that the original CNN was designed to classify, such that the resource requirement is proportionally scaled down?'' We use the term resource scalability to refer to this concept, and develop a methodology for automated synthesis of resource scalable CNNs from an existing optimized baseline CNN. The synthesized CNN has sufficient learning capacity for handling the given IoT application requirements, and yields competitive accuracy. The proposed approach is fast, and unlike the presently common practice of CNN design, does not require iterative rounds of training trial and error.
\end{abstract}
%
%
\maketitle
\section{Introduction}
\label{sec:introduction-dac}
Deep Convolutional Neural Networks (CNNs) have drastically extended our abilities in visual comprehension. 
Inspired by recent CNN-driven accomplishments, embedded platforms have witnessed a surge in the demand for visual recognition tasks. However, the functions and capabilities of these platforms differ from the powerful computation infrastructures that artificial intelligence practitioners use to host their models. The difference is twofold: first, embedded platforms are resource constrained. That is, they typically have a low-end processor, small memory footprint, limited storage, and a tight power budget. Second, applications of embedded platforms are usually mission-driven in that they are expected to only handle a few specific tasks. 
Hence, using a large CNN which is trained to recognize numerous object classes is neither efficient nor applicable in many embedded tasks. 

\begin{figure}
	\centering
	\includegraphics[width=6.5cm, height=4.7cm]{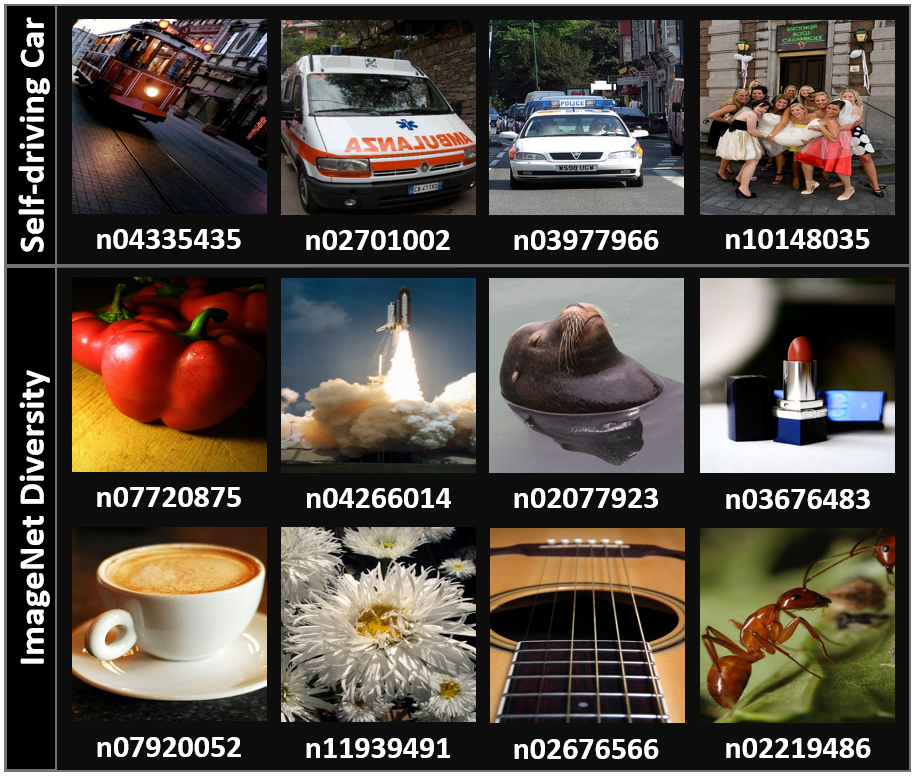}
	\caption{Sample images from ILSVRC~\cite{deng2009imagenet}. First row: images from classes that are relevant for a self-driving car. Second and third rows: images from just a few out of many classes that are not required for such an application.}
	\label{Fig:diversity}
\end{figure}

To illustrate this point, Figure \ref{Fig:diversity} presents images from a subset of classes available in ILSVRC dataset~\cite{deng2009imagenet}. The first row shows images from a few classes that are of interest for self-driving cars, whereas the other rows show samples from several classes that are not useful for this application.
Bridging the gap between embedded-system-grade deep learning requirements and server-grade research that develops such networks seems viable by redesigning a new network for each new embedded application. However, this approach faces its own challenges, such as the time-consuming nature of CNN design, which increases time to market, and difficult nature of developing competitive models for IoT practitioners. 



Designing a CNN is highly trial-and-error-based and demands many rounds of experimental training. That is, there does not exist a theoretical framework to guide CNN design  (distinct from its training) with guaranteed model accuracy. This begs the question: \emph{How can we leverage an existing optimized neural network model to automate the CNN design process for an IoT application whose objects of interest are a fraction of categories that the original CNN was designed to classify, such that the resource requirement is scaled down proportionally and the classification accuracy is competitive for the classes of interest to the application?}

Other than architecture design, creating an IoT-grade task-specific CNN is challenging due to two additional reasons: First, in a CNN, many kernels extract features that are beneficial in understanding all classes. Such feature extractors form and enrich by utilizing all images in the dataset. Designing a CNN for a subset of a dataset diminishes the number of training images which in turn deteriorates the quality of feature extraction. Second, lack of access to abundant training data makes a network more susceptible to overfitting. 

In this paper, we present a methodology for synthesis of resource-scalable CNNs, which addresses the aforementioned challenges. Our proposed approach utilizes an optimized CNN model as a baseline, and constructs a custom model whose parameter budget is scaled down in proportion to the complexity of the target task.
\section{Related Work}
The desire to develop CNNs that are fast, power efficient, and can accommodate embedded platforms has intensified~\cite{verhelst2017embedded} after Krizhevsky et al. work~\cite{krizhevsky2012imagenet}. We briefly review the related prior art that our work is inspired from.


\textbf{Network compression:} Most of the prior work on designing computation-efficient CNNs focus on reducing the size of a trained network in order to make the inference model smaller, faster, and less energy-intensive. Taking advantage of the fact that most CNNs have a considerable amount of redundant parameters, the research community has used quantization~\cite{han2015deep} and pruning~\cite{han2015learning} to perform network compression.

\textbf{Compact model design:} On another front, researchers work to design compact CNNs in the first place instead of compressing existing models~\cite{iandola2017small}.

While this work is inspired by the aforementioned endeavors, it addresses a substantially different question. We focus on the fact that most embedded platforms are mission-restricted, and thus, their scope of interest only includes a subset of classes from many categories for which, a complex CNN is designed. The objective of our work is orthogonal to prior work, as it takes as input a trained network, which could be compressed, pruned or otherwise optimized, and aims to leverage application-specificity (and orthogonal dimension to compression, pruning, etc.) to further optimize it. In our experiments, we use GoogLeNet, which is intrinsically compressed due to use of compress-expand layers, as the base network ($\Psi$) to demonstrate this point.

We propose a principled approach for CNN design automation by scaling down an optimized large CNN to tailor it to the classification requirements of a given IoT application. To the best of our knowledge, two recent attempts at solving the problem are reported: Distill-Net~\cite{DistillNet} and Octopus~\cite{Octopus}. Distill-Net profiles the inference to determine what parameters are extraneous for classes of interest to a given IoT application. Subsequently, it removes all such parameters. Octopus presents a new architecture in which each class has its own distinct parameter set. In such settings, if we remove a class, its corresponding parameters can be eliminated. In Section~\ref{sec:exp_rslt}, our work is compared with Distill-Net and Octopus. 
\section{Problem Statement}
\label{sec:problem_setup}
Let us assume a CNN, $\Psi$, is given, which yields state-of-the-art accuracy for classification of a set of  $\alpha$ classes, $A$, where $\alpha = |A|$.
Let us also assume that a particular IoT application requires recognizing a set of $\beta$ classes, $B$, such that $B \subset A$ where $\beta = |B|$ and $\beta \ll \alpha$.
\begin{equation}
\label{EQ:all_classes}
\small A = \{a_1, a_2, \cdots, a_\alpha\},~\small B = \{b_1, b_2, \cdots, b_\beta\}
\end{equation}
The goal is  to derive a new CNN, $\Psi'$, from $\Psi$ under the following conditions:\\
\textbf{Domain:} $\Psi'$, should be able to understand and classify images that belong to classes of set $B$. Hence, $\Psi'$ will have a $\beta$-way classifier instead of the $\alpha$-way classifier that exists in $\Psi$.\\
\textbf{Complexity:} The number of parameters in $\Psi'$ should proportionally scale down compared to that of $\Psi$. As discussed in Section~\ref{sec:Resource-Scalable CNN Synthesis Methodology}, such a scale down will in turn decrease the required processing resources on the target IoT platform.\\
\textbf{Admittance:} $\Psi'$ should be able to determine if an input image does not belong to the scope of interest of the target IoT application. That is, for an image which belongs to the set $\bar{B} =  A - B$, the output of $\Psi'$ should indicate that the input is not in the CNN's scope of expertise. Hence, $\Psi'$ shall refrain from classifying such an instance. This is an interesting notion and will be discussed thoroughly in Section~\ref{subsec:Scope-Aware Inference}.

In this paper, we use the phrase ``resource scalable CNN design'' to refer to the process of deriving $\Psi'$ from $\Psi$. Also, we use $\Gamma$ to refer to the average number of parameters that are used per class to extract distinguishing features that are necessary for its identification.


In a CNN with $\alpha$ classes, one can hope that \textbf{on average}, approximately $1/\alpha$ of all parameters are used for learning distinctive features to distinguish images that belong to each class. Hence, in an ideal setting, Equation~(\ref{eq:IdealGamma}) holds for $\Psi$ and Equation (\ref{Eq:Ideal}) yields a loose lower bound for the desirable number of parameters in $\Psi'$. In this equation, $\Phi'$ and $\Phi$ are the number of parameters in $\Psi'$ and $\Psi$, respectively. In this paper, the terms ``capacity of a network'' and ``number of parameters of a network'' are used interchangeably.
\begin{equation} 
\label{eq:IdealGamma}
\small \Gamma_{\Psi}^{Ideal} = \Phi/\alpha
\end{equation}
\begin{equation}
\label{Eq:Ideal}
\small \Phi' \ge \Phi \times \frac{\beta}{\alpha}
\end{equation}
Note that a considerable number of kernels are shared among different classes. While sharing makes it possible for each class to utilize more than $1/\alpha$ of all parameters, it decreases the number of parameters that are class-exclusive. That is, in practice, less than $1/\alpha$ of all parameters are used exclusively for each class. Hence, in deriving $\Psi'$, it is unrealistic to expect to achieve the ideal parameter budget. Nonetheless, it is beneficial to use this value as a lower bound to gauge the efficiency of a particular reduction from $\Psi$ to $\Psi'$.
\section{Resource-Scalable CNN Synthesis Methodology}
\label{sec:Resource-Scalable CNN Synthesis Methodology}
Performing only MAC operations, convolutional layers offer a small arithmetic intensity which makes them highly susceptible towards being memory/IO bounded\footnote{In the best scenario, \emph{if} we manage to keep all the kernels in cache and ignore the required memory transactions for write-backs, the arithmetic intensity will be equal to 2.}. On this ground, it is cogent to argue that the required computing resource for a platform targeted to host a CNN model should be proportional to the CNN's memory footprint. We can control a CNN's capacity and hence the processing resources that it demands by manipulating its kernel sizes. It is worth noting that decreasing the kernel sizes will in turn reduces the computational burden. In the rest of this section, we propose a methodology for scaling down a CNN in order to tailor it to the demands of an IoT-grade classification application.
\subsection{Generic CNN Architecture}
\begin{figure}
	\centering
	\includegraphics[width=\columnwidth]{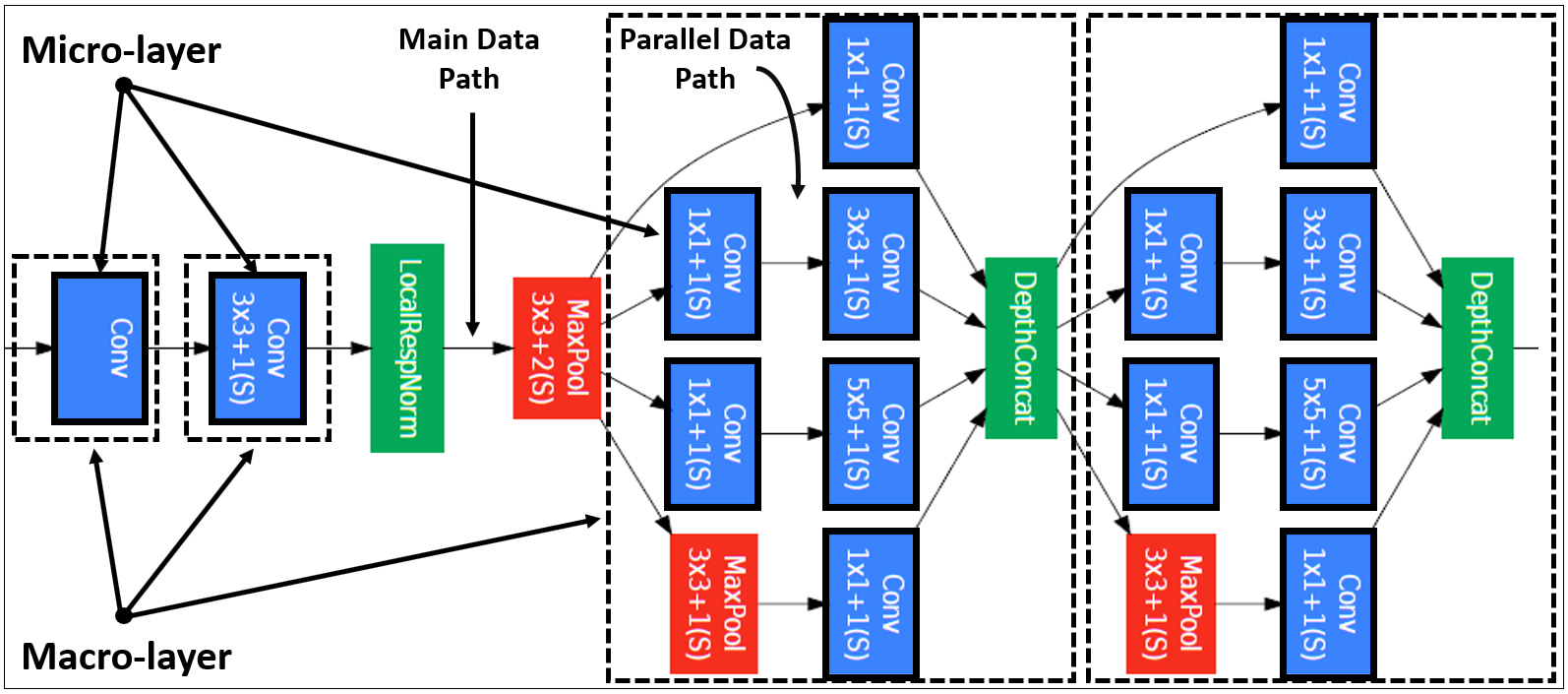}
	\caption{Macro-layers and Micro-layers in the Inception architecture~\cite{szegedy2015going}. CNNs generally include a sequence of macro-layers (shown with dashed black bounding boxes) and each macro-layer can have one or more micro-layers (shown with solid black bounding boxes). While macro-layers are necessarily sequential, micro-layers may operate in parallel or sequential fashion.}
	\label{fig:gennet}
\end{figure}
In all CNNs, data generally flows from the first feature map (i.e., the input) to the last feature map (i.e., the output) in a sequential fashion. That is, the input to each convolutional layer is the output of its predecessor layer. Even though the main data stream in a CNN is sequential, in modern neural networks, there are many local parallel data paths. The philosophy behind having such paths is to equip the network with a variety of kernels in order to enable it to extract features with different characteristics from the same input. Hence, the network will be less hindered by choice of hyper-parameters. Figure~\ref{fig:gennet} illustrates the Inception architecture~\cite{szegedy2015going}, a commonly-used CNN architectural blueprint, which is adopted by many prominent models such as SqueezeNet~\cite{iandola2016squeezenet}. CNNs' layers can generally belong to two different types. We use the terms ``micro-layer'' and ``macro-layer'' to refer to these layer types. Since the layer type will be very important in the rest of this paper, in what follows, we define each term clearly.\\
\textbf{Micro-layers:} We use the term micro-layer to refer to a convolutional unit that includes only one filter-bank (i.e., one set of kernels). Removing a micro-layer does not necessarily bisect the network into two separate parts. Micro-layers can be composed in parallel, and they cannot include any sub-layers.\\
\textbf{Macro-layers:} We use the term macro-layer to refer to a CNN layer that includes one or more micro-layers. Removing each macro-layer bisects the CNN into two separate parts. By definition, a CNN cannot have parallel macro-layers, since such a structure would simply be a larger macro-layer.\\
In deriving $\Psi'$ from $\Psi$, we preserve and scale (when applicable) non-convolutional layers such as max-pooling and batch-normalization. In the interest of simplicity, our discussions in this paper is only focused on convolutional layers where the major part of the computation happens.
\subsection{Model Scaling Principles}
To keep the discussion generic, we assume that members of $B$ are uniformly sampled from $A$ (Equations~(\ref{EQ:all_classes})) and they may include features that vary in scale and extraction complexity. Hence, we do not make any simplifying assumptions regarding the variety and complexity of kernels in $\Psi'$ compared to those of $\Psi$. On this ground, in driving $\Psi'$ from $\Psi$, we preserve the structural characteristic of the network, namely, kernels width and height, paddings, stride values, choice of pooling specifications, and regularizations. 

While our proposal does not include altering the variety and configuration of feature extractors in $\Psi$ and $\Psi'$, it does suggest scaling down the network capacity in $\Psi'$, since $|B| \ll |A|$. We introduce three knobs for CNN capacity control and indicate which one has the best performance in satisfying the requirements of our proposed method\footnote{This shall not be confused with kernel optimization approaches such as the use of asymmetric kernels. Our discussion targets this question: ``In case the capacity of a network is large for a task, how can we make it smaller?''}.\\
\textbf{(1) Number of Layers}: Decreasing the number of layers reduces the parameter budget in a model materially. In a CNN, feature representation vary across different layers. In deep layers the extracted features are semantically related to the input and they are distinctive enough to be used by the classifier to discriminate between different classes. As a result, removing a layer decreases the quality of features that are fed to the classifier. Hence, the accuracy is likely to drop.\\
\textbf{(2) Micro-layer Depth}: Depth of a micro-layer indicates the number of kernels of the same dimension that exist in the micro-layer. These replicas have the exact same dimension, the same input, and their outputs will be concatenated together to form a unique output for the micro-layer. Such a redundancy is required for a network to learn distinct features from different classes that are structurally analogous.\\
\textbf{(3) Micro-layer Breadth}: Micro-layer breadth is determined by the kernel width and height. Decreasing the kernel width or height diminishes receptive field of neurons which in turn makes it insufficient for capturing larger-scale features. Such a restriction deteriorates the quality of features extraction.

Since in the process of deriving $\Psi'$ from $\Psi$ we want to preserve the network's ability in extracting a wide variety of distinctive features, we neither change the number of layers nor the micro-layers breadth. What will be subject to change is the micro-layers depth. This is a cogent choice since $\Psi'$ is being designed for set $B$ where $|B| \ll |A|$.

Let us assume $\Psi$ has $N$ macro-layers $\{l_1, l_2, \cdots ,l_N\}$ each of which includes a set ($\mu$) of one or more micro-layers with different kernel dimensions. Assuming that a macro-layer $l_i$ includes $n_i$ micro-layers, Equation~(\ref{Eq:microlayers}) presents a concise form for showing dimensions of each of these micro-layers in $l_i$. In this equation, $w$, $h$, and $d$ are filter-banks width, height, and depth, respectively.
\begin{equation}
\label{Eq:microlayers}
\small \mu^{l_i} = \{(w^{l_i}_1, h^{l_i}_1, d^{l_i}_1), \cdots, (w^{l_i}_{n_i}, h^{l_i}_{n_i}, d^{l_i}_{n_i})\}
\end{equation}
We aim to manipulate micro-layers depth to control the parameter budget of a CNN. However, in order to preserve the characteristics of the optimized input CNN ($\Psi$) to a higher extent, we perform such manipulations under two guiding principles that are discussed in what follows:\\
\textbf{(1) Bottleneck Avoidance:} In general, as we move from shallower macro-layers towards deeper ones, the number of channels in the consecutive macro-layers increases. We want to preserve this quality while changing the depth of micro-layers. That is, Equation~(\ref{EQ:loose_ineq}) should hold for all layers of $\Psi'$.
\begin{equation}
\label{EQ:loose_ineq}
\scriptsize \sum_{j = 1}^{n_i} d^{l_i}_j \nll \sum_{j = 1}^{n_{(i - 1)}} d^{l_{(i - 1)}}_j
\end{equation}
\textbf{(2) Affine Scaling:} In deriving $\Psi'$ from $\Psi$, we would like to preserve the contribution ratio of each micro-layer to the feature maps generated in the corresponding macro-layer. To do so, we need to ensure that for each macro-layer $l_i$, Equation~({\ref{EQ:equal_ratios}}) holds. In this equation, parameter $q$ is depth of micro-layers in $\Psi'$. To be precise, $q^{l_i}_{x}$, is the depth of micro-layer $x$ in macro-layer $l_i$ of CNN $\Psi'$.
\begin{equation}
\label{EQ:equal_ratios}
\small \frac{d^{l_i}_1}{q^{l_i}_1} = 	\frac{d^{l_i}_2}{q^{l_i}_2} = \cdots = 	\frac{d^{l_i}_{n_i}}{q^{l_i}_{n_i}}
\end{equation}

Satisfying Equation~(\ref{EQ:equal_ratios}) requires us to scale all micro-layers of each macro-layer by a single factor. Since each channel depth after scaling must be an integer, this single factor has to be a common divisor of depth of all micro-layers in the macro-layer. Equation~(\ref{Eq:common-divisors}) simplifies this concept.
\begin{multline}
\label{Eq:common-divisors}
\small \text{Possible Scaling Factors}~(l_i) = \\ \{\text{Common Divisors}~(d^{l_i}_1, d^{l_i}_2, \cdots, d^{l_i}_{n_i})\}
\end{multline}
Hence, we reduce the process of deriving $\Psi'$ from $\Psi$ to the following problem: ``Given $\Phi$ and $\Phi'$ from Equation~(\ref{Eq:Ideal}) and a set of possible scaling factors for each macro-layer of a given CNN, what scaling factors should be selected to obtain the largest CNN with $\Phi'$ parameter budget, while avoiding bottleneck creation?''\footnote{In the next subsection, we further analyze and refine Equation~(\ref{Eq:Ideal}) to derive Equation~(\ref{Eq:NotIdeal}). Nonetheless, answer to this question remains an essential part of deriving $\Psi'$ from $\Psi$.} This problem maps to a variation of the Multiple-Choice (MC) knapsack problem, in which, the parameter budget of $\Phi'$ gives the knapsack size. Each macro-layer is a class in the MC-knapsack instance, and only one item from each class, i.e., one scaling factor from the list given by Equation~(\ref{Eq:common-divisors})), can be selected. The size and reward associated with an item are both equal to the scaled number of parameters in the macro-layer, if the associated scaling factor is selected.

Bottleneck avoidance, which does not exist in the standard MC-knapsack formulation, is our domain-specific additional constraint that needs to be met. Another domain-specific subtlety is that there exists dependency among rewards that are to be collected from scaling factors in adjacent classes. The reason is that selection of a specific scaling factor in a layer impacts the number of output feature maps of the layer, which in turn, influences the reward associated with scaling factors of the subsequent layer. Instances arising from practical CNNs have limited parameter budget of $\Phi'$, and only several choices of scaling factor in each layer. Thus, despite NP-completeness of the general problem, practical instances can be solved optimally in a reasonable time using dynamic programming.

\subsection{An Illustrative Example}
\begin{table}
	\caption{Specifications of convolutional layers in AlexNet~\cite{krizhevsky2012imagenet}. IFMs and OFMs stand for Input Feature Maps, and Output Feature Maps.}
	\label{tbl:alex_compress}
	\scalebox{0.68}{
		\centering
		\begin{tabular}{ccccccc}
			\toprule[0.05cm]
			\textbf{Layer Name} &  \textbf{\# IFMs}  &  \textbf{\# OFMs}   &  \textbf{\# Rows}  & \textbf{\# Cols.}  & \textbf{Kernel size}  & \textbf{\# Params. (M)} \\ \toprule[0.05cm]
			 Conv \#1  &  3  & 96  & 55 & 55 & 11 &      0.03      \\
			 Conv \#2  & 96  & 256 & 27 & 27 & 5  &      0.61      \\
			 Conv \#3  & 256 & 384 & 13 & 13 & 3  &      0.88      \\
			 Conv \#4  & 384 & 384 & 13 & 13 & 3  &      1.33      \\
			 Conv \#5  & 384 & 256 & 13 & 13 & 3  &      0.88      \\ \bottomrule[0.05cm]
		\end{tabular}
	}
\end{table}
In this section, we use AlexNet~\cite{krizhevsky2012imagenet} to demonstrate the idea. This CNN is used strictly due to its simplicity and popularity. It is worth noting that AlexNet has an outdated architecture that includes high-dimensional fully connected layers, and thus, it is not a realistic target of our research.

Table~\ref{tbl:alex_compress} shows the specifications of convolutional layers in AlexNet~\cite{krizhevsky2012imagenet}. Let us assume for a given IoT application only 8\% of AlexNet's capacity is required. Using the proposed method, this network can be scaled down in 228832 different ways\footnote{For comparison, note that this number for an industry-strength CNN such as GoogLeNet equals to $6.39 \times 10^{10}$ with affine scaling and $1.72 \times 10^{54}$ without it. Both numbers include solutions that ignore the bottleneck constraint, and they are derived using an approach analogous to Equation~(\ref{eq:possible_solutions}).}, as it is detailed in  Equation~(\ref{eq:possible_solutions}). In this equation, the function $d\left(.\right)$ returns the number of divisors of its input. Equation~(\ref{eq:possible_solutions}) yields all scale-down possibilities, disregarding the bottleneck avoidance constraint. Among all possible scale-down factors, 2335 solutions offer a capacity of $\Phi'$ in the acceptable threshold of 0.002, such that $|\frac{0.08\Phi - \Phi'}{\Phi}| < 0.002$.
\begin{equation}
\label{eq:possible_solutions}
\small \text{\#Solutions} = d\left( 96\right) \times d\left(256\right) \times d\left(384\right)\times d\left(384\right)\times d\left(256\right)
\end{equation}
Table~\ref{tbl:alex_solutions} presents 4 sample answers for this example. While all of them offer a desirable capacity reduction, only one (\#4) meets our proposed constraints. To further elaborate, solution \#1 introduces a severe bottleneck in layer 3, \#2 creates bottleneck in layer 4, and solution \#3 adds bottlenecks to layers 3 and 5. The problem of finding an efficient answer among a large pool of potential solutions is challenging for state-of-the-art deep CNNs.
\begin{table}
	\caption{Example layer-wise scaling scenarios aiming to reduce 92\% of AlexNet's convolutional capacity.}
	\label{tbl:alex_solutions}
	\scalebox{0.58}{
		\centering
		\begin{tabular}{ccccccccccccc}
			\toprule[0.06cm]
			\multirow{2}{*}{\textbf{\#}}&\multicolumn{5}{c}{\textbf{Scale Factors}} && \multicolumn{5}{c}{\textbf{Number of OFMs}}\\\cmidrule[0.04cm]{2-6}\cmidrule[0.04cm]{8-12}
			&\textbf{Con.1} &  \textbf{Con.2} & \textbf{Con.3} & \textbf{Con.4} & \textbf{Con.5} && \textbf{OFMs1} & \textbf{OFMs2} & \textbf{OFMs3} & \textbf{OFMs4} & \textbf{OFMs5} \\ \toprule[0.06cm]
			Baseline & 1 & 1 & 1 & 1 & 1 && 96 & 256 & 384 & 384 & 256 \\
			Case \#1 & 24 & 4 & 128 & 3  &  1  &&  4 & 64 & 3 & 128 & 256 \\
			Case \#2 & 2  & 4 &  2  & 12 &  2  &&  48 & 64 & 192 & 32 & 128\\
			Case \#3 & 3  & 1 & 12  & 8  & 128 &&  32 & 256 & 32 & 48 & 2\\
			Case \#4 & 8  & 8 &  4  & 3  &  2  && 12 & 32 & 96 & 128 & 128 \\ \bottomrule[0.06cm]
		\end{tabular}
	}
\end{table}
\subsection{Scope-Aware Inference}
\label{subsec:Scope-Aware Inference}
Conventional CNNs assign any input image into one of their output classes, even if it belongs to none of them. As a case in point, AlexNet~\cite{krizhevsky2012imagenet} classifies a picture of red blood cells to honeycomb. It is very desirable if a CNN can manage to determine whether an input image belongs to its scope of expertise. If so, the CNN should attempt to classify it. Otherwise, the classification task should be rejected.

In this section, we address this issue by adding a miscellaneous class to $\Psi'$. Classifying an input image into this class indicates that the CNN believes the input is not in its scope of expertise. We use the dataset that corresponds to $\bar{B}$ to train the miscellaneous class.

It is challenging to realize such a screening mechanism since a CNN must have a reasonable understanding of an input image in order to determine whether or not it belongs to its scope of expertise. As a result, the neural network must include a vast variety of additional kernels for recognition of members of set $\bar{B}$. In a complicated dataset such as ImageNet, a high resemblance may exist between images from different classes as illustrated in Figure~\ref{fig:sai_hard}. Hence, since $B$ is randomly selected, its members can be highly analogous to those of the miscellaneous class. In such settings, scope-aware inference will be further challenging since the CNN must learn to extract a plethora of coarse- and fine-grained features that are distinctive enough for discrimination between classes of $B$ and $\bar{B}$ with high correlations. We hypothesize that the required parameter budget for learning the miscellaneous class is a semi-linear function of $|\bar{B}|$. Learning to categorize members of $|\bar{B}|$ in a single miscellaneous class is easier than learning to distinguish them from each other. A complete differentiation requires a parameter budget of $\Phi \times |\bar{B}|/\alpha$ for each class. Hence, for gathering all members of $\bar{B}$ in the miscellaneous class, we expect to need a parameter budget of $\lambda \times \Phi \times |\bar{B}|/\alpha$ where $\lambda < 1$.
As a result, the total parameter budget for $\Psi'$, first introduced in Equation~(\ref{Eq:Ideal}), is more accurately lower bounded by:
\begin{figure}
	\centering
	\includegraphics[width=\columnwidth]{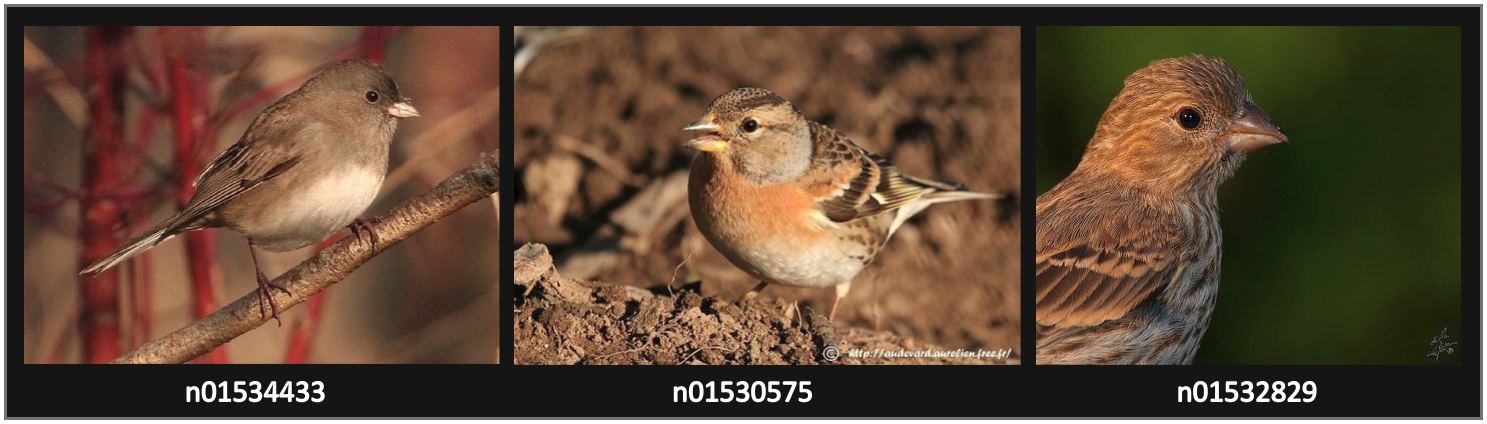}
	\caption{Example images of three distinct, albeit very similar, classes in ImageNet. Since set $B$ is randomly selected from $A$, its members may be very analogous to those of $\bar{B}$, considerably complicating the task of $\Psi'$.}
	\label{fig:sai_hard}
\end{figure}
\begin{equation}
\label{Eq:NotIdeal}
\small \Phi' \ge (\Phi \times \frac{\beta}{\alpha}) + (\lambda \times \Phi \times \frac{\alpha - \beta}{\alpha})
\end{equation} 
\subsection{Class-imbalance Resolution}
Since $|B| \ll |A|$, then $|B| \ll |\bar{B}|$. As a result, in training a scope-aware CNN, a very large number of input images in every batch would belong to set $\bar{B}$. In other words, a large portion of training examples would represent the miscellaneous class. Such a high imbalance in the input data can easily overwhelm the classifier and bias it towards learning the miscellaneous class only.

In a classification problem, the ratio of classes should be approximately the same in order to ensure that different classes \emph{almost} have an equal share in the computed gradient. This is a critical point since the value of the gradient determines how weights change in every round of backpropagation. That is, the value of gradient dictates what a model learns. To address the class-imbalance issue, we simply need to select the same number of training instances for each class. Hence, even though the dataset includes many images for the miscellaneous class, we need to pick a certain number of them in every iteration.

\begin{figure}
	\centering
	\includegraphics[width=\columnwidth]{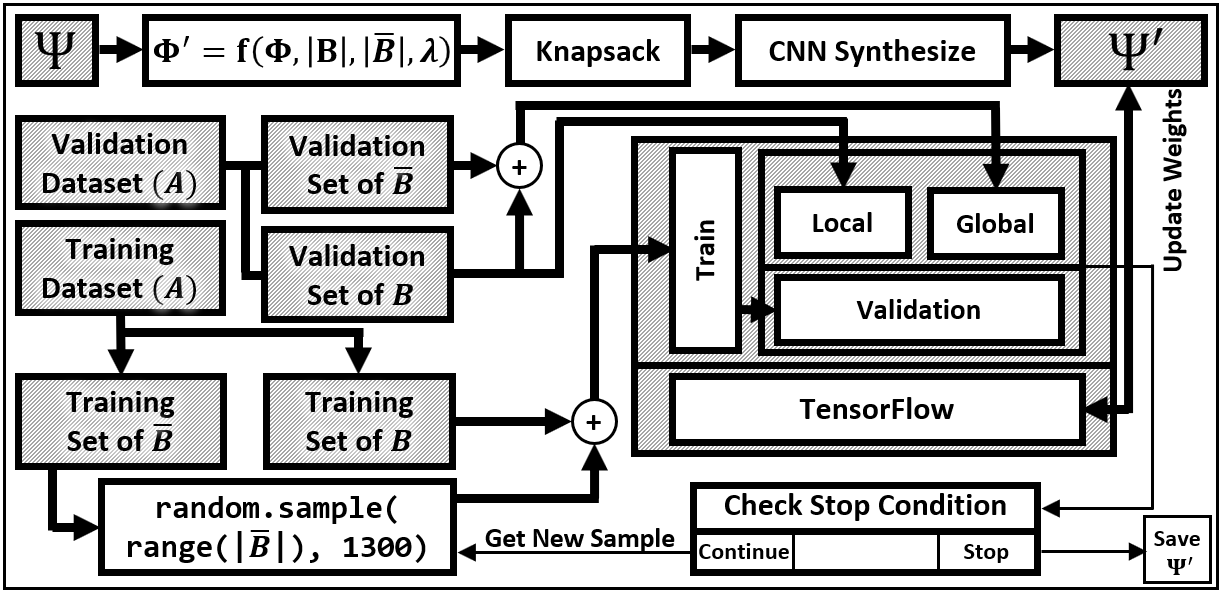}
	\caption{The process of CNN synthesis for a given IoT-grade classification task. The proposed approach uses an optimized CNN as a baseline and automatically scales it down to fulfill the classification demands of the given task. The synthesized CNN, $\Psi'$, performs high-accuracy scope-aware inference, at a fraction of computational cost of the given's CNN.}
	\label{fig:psi_to_psi_prime}
\end{figure}

We resolved the class-imbalance issue by implementing a data generator that randomly selects 1300 new images from all classes of $\bar{B}$ in each iteration. The number 1300 is the size of training set for each class of ImageNet. Note that training images for classes of $B$ remain the same in different iterations. The methodology is summarized in Figure~\ref{fig:psi_to_psi_prime}. In this figure, function $f$ computes Equation~(\ref{Eq:NotIdeal}).
\subsection{A Remedy for Overfitting} 
\begin{table*}
	\caption{Performance comparison between different approaches for resource-scalable CNN design targeting IoT-grade applications.}
	\label{tbl:accuracy_cnns}
	\scalebox{0.57}{
		\begin{tabular}{cccccccccccacccacccccccccc}
			\toprule[0.05cm]
			\multirow{2}{*}{\makecell{\# Cls. in Target\\ IoT-grade CNN}} & \multicolumn{3}{c}{\textbf{Number of Parameters (M)}} & & \multicolumn{3}{c}{\textbf{GFLOPS}} & & \multicolumn{3}{c}{\textbf{Local Accuracy (\%)}} & & \multicolumn{3}{c}{\textbf{Global Accuracy (\%)}}& & \multicolumn{3}{c}{\textbf{Inference Time (ms)}} && \multicolumn{3}{c}{\textbf{Speedup to GoogLeNet}}\\\cmidrule[0.04cm]{2-4}\cmidrule[0.04cm]{6-8}\cmidrule[0.04cm]{10-12} \cmidrule[0.04cm]{14-16} \cmidrule[0.04cm]{18-20} \cmidrule[0.04cm]{22-24}
			& Distil. & Octopus & Proposed & & Distil. & Octopus & Proposed & & Distil. & Octopus & Proposed & & Distil. & Octopus & Proposed && Distil. & Octopus & Proposed && Distil. & Octopus & Proposed\\\toprule[0.05cm]
			5  &2.61&0.18& 1.77 &&2.20& 0.12 & 1.39 && 68 & 68 & \textbf{85} & & 0.34 & 0.34 & \textbf{94.01} & & 990  & 68  & 671 & & 2.68X & 38.83X & 3.95X\\
			10 &3.58&0.18& 1.80 &&2.42& 0.12 & 1.40 && 68 & 68 & \textbf{83} & & 0.68 & 0.68 & \textbf{87.12} & & 1358 & 68  & 683 & & 1.95X & 38.83X & 3.88X\\
			15 &3.97&0.18& 1.82 &&2.56& 0.12 & 1.42 && 68 & 68 & \textbf{84} & & 1.02 & 1.02 & \textbf{88.33} & & 1506 & 68  & 690 & & 1.76X & 38.83X & 3.84X\\
			20 &4.55&0.18& 1.85 &&2.78& 0.12 & 1.46 && 68 & 68 & \textbf{81} & & 1.36 & 1.36 & \textbf{86.00} & & 1726 & 68  & 702 & & 1.54X & 38.83X & 3.78X\\
			25 &4.90&0.24& 1.88 &&2.89& 0.18 & 1.47 && 68 & 68 & \textbf{82} & & 1.70 & 1.70 & \textbf{83.59} & & 1858 & 91  & 713 & & 1.43X & 29.12X & 3.72X\\
			30 &5.02&0.36& 1.90 &&2.92& 0.19 & 1.49 && 68 & 68 & \textbf{84} & & 2.04 & 2.04 & \textbf{84.02} & & 1904 & 137 & 721 & & 1.39X & 19.42X & 3.68X\\\toprule[0.04cm] 
			Average &4.11&0.22& 1.84 &&2.63& 0.14 & 1.44 && 68 & 68 & \textbf{83.17}&&1.19 & 1.19 & \textbf{87.18} & & 1557 &83.33&696.67&& 1.79X & 33.96X & 3.81X\\ 
			\bottomrule[0.05cm]
		\end{tabular}
	}
\end{table*}
\begin{table}
	\caption{Performance of synthesized CNNs vs. baseline CNN.}
	\label{tbl:accuracy_cnns_compared_to_base}
	\scalebox{0.65}{
		\centering
		\begin{tabular}{cccccccc}
			\toprule[0.05cm]
			\multirow{2}{*}{Name}  &\multirow{2}{*}{\rotatebox{0}{\makecell{No. of\\ Classes}}} & \multirow{2}{*}{\makecell{Parameter Ratio\\to GoogLeNet}} & \multicolumn{2}{c}{Global Accuracy}&& \multicolumn{2}{c}{Local Accuracy} \\\cmidrule[0.04cm]{4-5} \cmidrule[0.04cm]{7-8} 
			& && GoogLeNet & Proposed                 && GoogLeNet     &Proposed             \\\toprule[0.05cm]
			$\Psi'_1$ & 5  & 0.25 & 0.99 & 0.94 &  & 0.68 & 0.85 \\
			$\Psi'_2$ & 10 & 0.26 & 0.99 & 0.87 &  & 0.68 & 0.83 \\
			$\Psi'_3$ & 15 & 0.26 & 0.99 & 0.88 &  & 0.68 & 0.84 \\
			$\Psi'_4$ & 20 & 0.27 & 0.99 & 0.86 &  & 0.68 & 0.81 \\
			$\Psi'_5$ & 25 & 0.27 & 0.98 & 0.83 &  & 0.68 & 0.82 \\
			$\Psi'_6$ & 30 & 0.27 & 0.97 & 0.84 &  & 0.68 & 0.84 \\\bottomrule[0.05cm]
		\end{tabular}
	}
\end{table}
The miscellaneous class needs to learn many different images from diverse categories. Hence, it has a much higher learning complexity compared to other classes. As a result, a CNN learns the normal classes and starts getting overfitted on them long before it learns the miscellaneous class. In our settings, the main reason for such a phenomenon is the choice of loss: categorical cross entropy. Categorical cross entropy, which is widely used in image classification problems, is a greedy loss. That is, it may still heavily impact the gradient towards learning instances that are already learned. Such an excessive learning leads to overfitting in our settings.

Since classes in $\Psi'$ vary in learning complexity, we need to find a mechanism that attenuates the impact of learned classes on the gradient. The benefit is twofold: First, the network will not overfit on learned classes. Second, unlearned instances will have a higher impact on the gradient, making the network more inclined towards learning them. To achieve this, we need to utilize a loss function that slows down the training process for learned instances. In particular, we adopt focal loss, a non-greedy loss function fulfilling our requirements, which is recently introduced for object detection (not classification)~\cite{lin2018focal}. Use of focal loss in the pure classification context can be beneficial if $B \nequiv A$, and to the best of our knowledge, this work for the first time utilizes focal loss in the image classification context.
\section{Results and Discussions}
\label{sec:exp_rslt}
\subsection{Performance Metrics}
The notion of scope-aware inference enables a CNN to perform two nested classification tasks. Hence, it requires two metrics to gauge the performance of each. The first metric measures the classification accuracy on the entire validation set of ImageNet, and the second one measures it on the validation set of classes that are in the scope of interest of our application. We refer to the first and second metrics using the terms global and local accuracy, respectively.
Measuring the global accuracy is required to determine the efficacy of scope-aware inference, whereas measuring the local accuracy is essential to determine the qualify of recognition for classes that are within our scope of interest. In the absence of local accuracy, a network might learn to classify every single input as a member of the miscellaneous class (i.e., does not learn anything) and still yields a global accuracy of 99.9\% (since, $|B| \ll |A|$). Likewise, in the absence of global accuracy, a CNN will not be able to determine what instances do not belong to its scope of expertise.
\subsection{Studying $\Psi'$ when $\Psi =$ GoogLeNet}
We used Distill-Net~\cite{DistillNet}, Octopus~\cite{Octopus}, and the proposed approach to derive CNNs for 6 different IoT-grade tasks while using GoogLeNet~\cite{szegedy2015going} as our baseline neural network, $\Psi$. In all experiments, members of $B$ are randomly selected. We used the ImageNet 2012 competition training dataset~\cite{russakovsky2015imagenet} for the training process and the derived CNNs are tested on the validation data from the same dataset. 

The global and local accuracy measurements of the derived CNNs are presented in Table~\ref{tbl:accuracy_cnns_compared_to_base}. In our experiments we empirically found the value of $\lambda$ to be 0.25. With a considerable parameter reduction, all instances of $\Psi'$ yield a much higher local accuracy compared to the base network (i.e., GoogLeNet) while yielding a reasonably high global accuracy. In addition, the CNNs are designed with a scope-aware-inference philosophy in mind. Hence, they can determine, with a high confidence (84\% - 94\%), if an input image does not belong to their scope of expertise. It is also worth noting that the proposed mechanism for CNN synthesis is fully automated.

Table~\ref{tbl:accuracy_cnns} compares the performance of CNNs that are synthesized using Distill-Net, Octopus, and the proposed methodology. Octopus-based CNNs yield the best parameter budget and have the smallest computational burden. This is expected since they are not designed to support scope-aware inference. Hence, their network capacity will be proportional to $|B|$. A scope-aware CNN requires to have a comprehensive understanding of members of $|B|$ an a reasonable perception of members of $|\bar{B}|$. Hence, the parameter budget for a such a network would be proportional to $|B| + \lambda|\bar{B}|$ which is much larger than $|B|$. Coefficient $\lambda < 1$ moderates the impact of $|\bar{B}|$ in the parameter budget. It is worth noting that even though Distill-Net-based CNNs do not offer scope-aware inference, their parameter and computational budget is inferior to those of the CNNs designed using the proposed methodology. The execution times for the proposed CNNs is estimated using the model developed in~\cite{motamedi2018cappuccino} targeting Qualcomm Snapdragon 800 SoC, and the results are shown in Table~\ref{tbl:accuracy_cnns}.

CNNs synthesized using the proposed approach outperform Distill-Net-based and Octopus-based neural networks in terms of both local and global accuracy. Our CNN synthesis mechanism achieves a local accuracy of 83.17\% on average which is superior to 68\% accuracy that other schemes offer. It also yields a considerably high global accuracy of 87.18\%. The lack of support for scope-aware inference in Distill-Net and Octopus makes them unsuccessful in achieving a competitive global accuracy.


\section{Conclusion}
In this paper, we proposed a resource-scalable CNN design methodology that can be used to eliminate extraneous classes in a CNN which are not required for a particular embedded application. The proposed solution is fully automated and can be used by a machine to approximate the required CNN for a given IoT-grade task. Our experimental results show that the synthesized CNNs can yield state-of-the-art local and global accuracy on embedded-grade classification tasks.
\bibliographystyle{unsrt}
\tiny{\bibliography{arxiv}}
\end{document}